\documentclass[runningheads]{llncs}
\usepackage[width=122mm,left=12mm,paperwidth=146mm,height=193mm,top=12mm,paperheight=217mm]{geometry}
\usepackage[T1]{fontenc}
\usepackage{graphicx}
\usepackage{graphicx}
\usepackage{enumitem}
\usepackage{xspace}
\usepackage{subfiles}
\usepackage{subcaption}
\usepackage{contour}
\usepackage{booktabs}
\usepackage[normalem]{ulem}
\usepackage{multirow}
\usepackage{eucal}
\usepackage{amsmath}
\usepackage{wrapfig}
\usepackage{pifont}
\usepackage[misc]{ifsym}

\usepackage[ruled, lined, linesnumbered, commentsnumbered, longend]{algorithm2e}

\usepackage{enumitem}
\usepackage{xcolor, colortbl}
\definecolor{jongha}{rgb}{0.5, 0.2, 0.94}

\newcommand{\myuline}[1]{%
  \uline{\phantom{#1}}%
  \llap{\contour{white}{#1}}%
}
\newcommand{\smallsection}[1]{{\noindent {\textbf{\myuline{#1}}}}}

\newcommand{\gdc}{\textsc{GDC}\xspace}
\newcommand{\svd}{\textsc{SVD}\xspace}
\newcommand{\jaccard}{\textsc{Jaccard}\xspace}
\newcommand{\adamic}{\textsc{A.A.}\xspace}
\newcommand{\tiara}{\textsc{Tiara}\xspace}
\newcommand{\leo}{\textsc{LEO}\xspace}

\newcommand{\algname}{\textsc{TiGer}\xspace}

\newcommand{\vecb}{\boldsymbol{b}}

\newcommand{\vech}{\boldsymbol{h}}

\newcommand{\vecl}{\boldsymbol{l}}

\newcommand{\vecz}{\boldsymbol{z}}

\newcommand{\cmark}{\ding{51}}
\newcommand{\xmark}{\ding{55}}
\begin{document}
\title{TiGer: Self-Supervised Purification for Time-evolving Graphs}

\author{Hyeonsoo Jo\inst{1}\orcidID{0000-0002-9281-8672} \and
Jongha Lee\inst{1}\orcidID{0000-0001-7197-3529} \and
Fanchen Bu\inst{2}\orcidID{0000-0003-0497-3902} \and
Kijung Shin\inst{1}\orcidID{0000-0002-2872-1526}\textsuperscript{(\Letter)}}

\authorrunning{H. Jo et al.}

\institute{Kim Jaechul Graduate School of AI, KAIST, Seoul, Republic of Korea \and School of Electrical Engineering,
KAIST, Daejeon, Republic of Korea \\
\email{\{hsjo,jhsk777,boqvezen97,kijungs\}@kaist.ac.kr}}

\maketitle              

\begin{abstract}

Time-evolving graphs, such as social and citation networks, often contain noise that distorts structural and temporal patterns, adversely affecting downstream tasks, such as node classification.
Existing purification methods focus on static graphs, limiting their ability to account for critical temporal dependencies in dynamic graphs.
In this work, we propose \algname (\textbf{\underline{\textsc{Ti}}}me-evolving \textbf{\underline{\textsc{G}}}raph purifi\textbf{\underline{\textsc{er}}}), a 
self-supervised method explicitly designed for time-evolving graphs. 
\algname assigns two different sub-scores to edges using (1) self-attention for capturing long-term contextual patterns shaped by both adjacent and distant past events of varying significance and (2) statistical distance measures for detecting inconsistency over a short-term period.
These sub-scores are used to identify and filter out suspicious (i.e., noise-like) edges through an ensemble strategy, ensuring robustness without requiring noise labels.
Our experiments on five real-world datasets show \algname filters out noise with up to 10.2\% higher accuracy and improves node classification performance by up to 5.3\%, compared to state-of-the-art methods.

\keywords{Graph Purification  \and Dynamic Graphs \and Robust Learning.}
\end{abstract}

\section{Introduction}\label{sec:intro}
Many real-world systems evolve over time, such as social networks and citation networks, resulting in a need for efficient methods to process and analyze these changes.
Such systems are commonly modeled as \textit{time-evolving graphs}.

\textit{Noise} is a prevalent challenge in time-evolving graphs~\cite{kang2019robust,li2024robust,zhang2023rdgsl}.
For instance, social networks may contain noisy interactions from spam accounts or erroneous users, distorting graph structures.
As graphs grow in scale, noise increases correspondingly, amplifying its adverse impact on downstream task performance and demanding robust methods to mitigate it effectively.

\textit{Graph purification} enhances the quality of graph data by detecting and mitigating noise that deviates from underlying \textit{patterns} (i.e., normal behaviors).
It typically measures a similarity of node pairs that form edges in the graph, identifying edges with low similarity as noise.
Existing methods for measuring similarity can be broadly categorized into local similarity-based and global similarity-based approaches, depending on the scope of the structural information they utilize.
Local similarity-based methods, such as the Jaccard Coefficient~\cite{sathre2022edge} and the Adamic-Adar Index~\cite{tian2020exploiting}, evaluate node-pair similarity based on local neighborhood information.
Global similarity-based approaches, such as singular value decomposition (SVD),
leverage global structural properties to remove noise-like edges. 
Despite their effectiveness, these techniques are designed for \textit{static graphs}, where only \textit{structure patterns} are considered.


To the best of our knowledge, we are the first to consider graph purification on time-evolving graphs.
This task presents unique challenges due to the dynamic nature of these graphs, where \textit{temporal patterns} play a critical role in distinguishing noise from meaningful connections~\cite{barros2021survey,kazemi2020representation,skarding2021foundations}.
Specifically, each time-evolving graph can be seen as a sequence of static snapshots with temporal dependencies between each other.
We can use graph purification methods for static graphs on different snapshots, but such a naive way treats different snapshots as independent and overlooks temporal patterns.
This limitation highlights the need for a graph purification method explicitly considering temporal patterns.

In this work, we propose \algname (\textbf{\underline{\textsc{Ti}}}me-evolving \textbf{\underline{\textsc{G}}}raph purifi\textbf{\underline{\textsc{er}}}),
a self-supervised graph-purification method explicitly designed for time-evolving graphs.
\algname comprehensively captures two different categories of temporal patterns that are common in real-world time-evolving graphs:
(1) \textit{long-term patterns} (e.g., contextual patterns shaped by both adjacent and distant past events of varying significance) 
and 
(2) \textit{short-term patterns} (e.g., consistency across consecutive time steps over a short period).
Specifically, two different sub-scores are assigned to edges using (1) self-attention for capturing long-term patterns and (2) statistical distance measures for detecting unusual deviations from short-term patterns, respectively.
Those two sub-scores are combined via an ensemble strategy, together with a third proximity-based sub-score, to ensure robust and efficient purification across time steps, all without noise labels.

Our contributions are summarized as follows:
\begin{itemize}
    \item \textbf{Novel Problem}: To the best of our knowledge, we are the first to consider graph purification on time-evolving graphs, where unique challenges of incorporating temporal patterns are involved.
    \item \textbf{Effective Method}: We propose \algname, a self-supervised purification method for time-evolving graphs. \algname leverages long-term and short-term temporal patterns to effectively identify noise, without requiring label supervision.
    \item \textbf{Empirical Validation}: Experiments on five real-world datasets show that \algname filters out noise with up to 10.2\% higher accuracy and achieves 5.3\% higher node classification accuracy, compared to state-of-the-art methods.
\end{itemize}
For \textbf{reproducibility}, we make our code and datasets publicly available at \cite{code}.

\section{Preliminaries and Related Work}\label{sec:prelim}

\subsection{Background}
\label{sec:prelim:backgound}

A \textit{static graph} $G = (V, E)$ is defined by its node set $V$ and edge set $E \subseteq \{(v_i, v_j) : v_i, v_j \in V\}$.
A \textit{time-evolving graph} can be represented as a sequence of static graphs $G^{(t)} = (V^{(t)}, E^{(t)})$ across time steps $t\in\{1,\cdots, T\}$, i.e., $\mathcal{G} = \{G^{(1)}, G^{(2)}, \cdots, G^{(T)}\}$.
In this work, a node or edge that appears at time step $t$ is treated as present in all subsequent time steps, i.e., $V^{(t)} \subseteq V^{(t')}$ and $E^{(t)} \subseteq E^{(t')}$, for any $t < t'$.
Let $\Delta{G}^{(t)} = (\Delta{V}^{(t)}, \Delta{E}^{(t)})$ denote the change at time step $t$ with new nodes
$\Delta{V}^{(t)} = V^{(t)} \setminus V^{(t-1)}$ and new edges
$\Delta{E}^{(t)} = E^{(t)} \setminus E^{(t-1)}$.


\subsection{Graph Purification}
\label{sec:prelim:purification}


Graph purification, a common pre-processing step in graph analysis, removes noisy edges from graph data by measuring a similarity between adjacent nodes~\cite{luo2021learning,shen2024graph,spinelli2021fairdrop,xu2022ned,entezari2020all}. 
Similarity measurement can be \textit{local}, using metrics like the Jaccard Coefficient~\cite{sathre2022edge} and Adamic-Adar Index~\cite{tian2020exploiting}, or \textit{global}, considering the entire graph structure, e.g., through singular value decomposition (SVD)~\cite{entezari2020all}.
Although effective, most graph purification methods are designed for static graphs and overlook temporal patterns crucial for identifying noise.



A related approach, graph augmentation, adds edges between similar nodes based on positions~\cite{klicpera2019diffusion,lee2023time} or labels~\cite{chen2020measuring} in both static and dynamic graphs. 
Recent work~\cite{lee2023time} models temporal dependencies through a weighted summation of high-order node proximities over time. In our experiments, we adapt such augmentation methods for purification by removing edges between dissimilar nodes.


\subsection{Temporal Patterns}
\label{sec:prelim:Temporal}

\textit{Temporal patterns} in time-evolving graphs are classified into long-term (contextual) and short-term (consistency) patterns.

\textit{Long-term patterns} capture temporal dependencies over extended periods, shaped by both recent and distant past events.
For example, in a coauthorship network, these patterns may reveal ongoing collaborations within a research group over several years.
Dynamic graph neural networks (DGNNs)~\cite{pareja2020evolvegcn,sankar2020dysat,tgn_icml_grl2020} capture such patterns using modules like RNNs or self-attention~\cite{vaswani2017attention}.

In contrast, \textit{short-term patterns} reflect consistent behaviors over a short period.
In social networks, users often interact with the same neighbors over short periods, showing consistency in interaction patterns.
Many DGNN studies~\cite{lee2024slade,tian2021self,yu2017link,zhu2022dynamic} exploit this consistency, assuming that node representations evolve gradually with minimal short-term changes.

\section{Proposed Method: \algname}
\label{sec:method}
In this section, we introduce our proposed method, \algname (\textbf{\underline{\textsc{Ti}}}me-evolving \textbf{\underline{\textsc{G}}}raph purifi\textbf{\underline{\textsc{er}}}), for dynamic graph purification.
\algname uses both long-term and short-term patterns to assign edge scores without ground-truth noise labels.
It consists of two dedicated modules:
(Module 1; Sec.~\ref{sec:method:long_module}) long-term module $M_{L}$ with self-attention to capture contextual patterns over long periods, and
(Module 2; Sec.~\ref{sec:method:short_module}) 
 short-term module $M_S$ that computes statistical distances to detect deviations from consistency over short periods.
\algname then ensembles the sub-scores from both modules to obtain the final purification scores (Sec.~\ref{sec:method:ensemble}), where an extra proximity-based sub-score is used to enhance robustness.

\subsection{Overview}

\begin{figure}[t]
    \centering
    \includegraphics[width=0.95\linewidth]{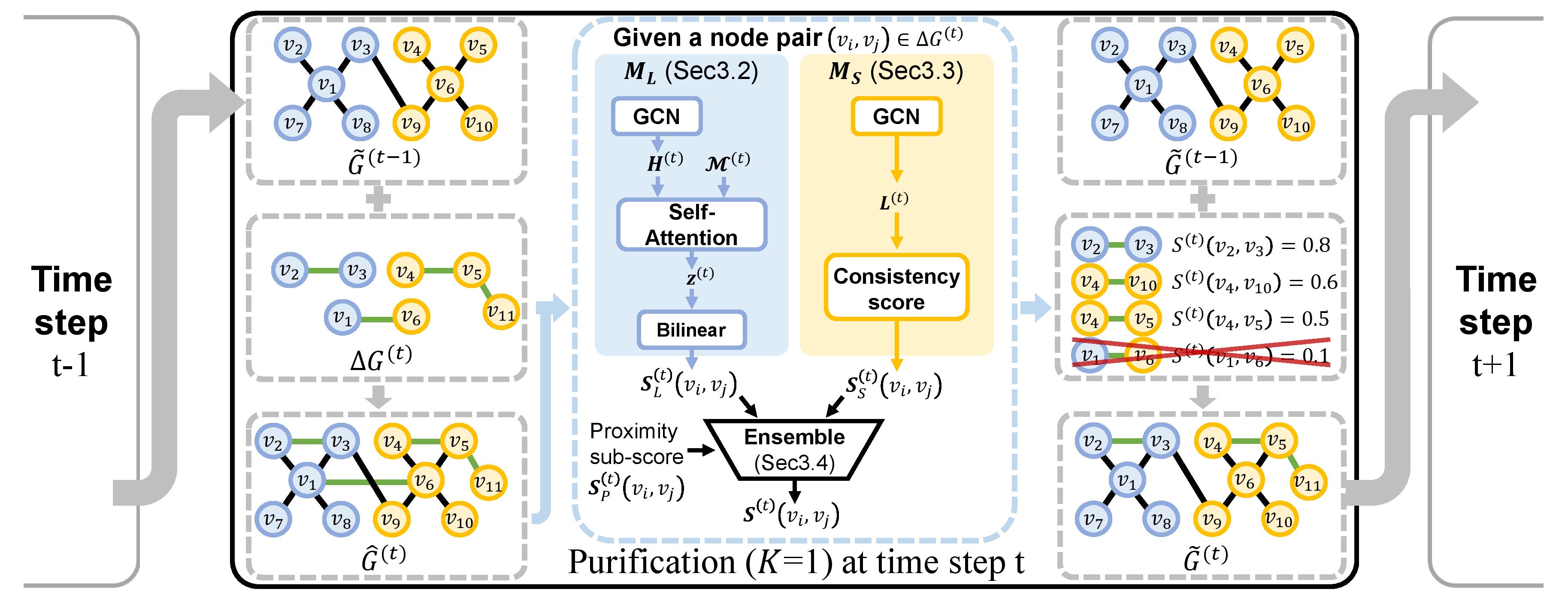} 
    \caption{Overall process of
    graph purification using \algname at each time step $t$.
    }
    \label{fig:prelim:purification}
\end{figure}

As illustrated in Fig.~\ref{fig:prelim:purification},
the overall process of graph purification on time-evolving graphs using \algname is iterative on each time step.
At each time step $t$, let $\tilde{G}^{(t - 1)}$ denote the \textit{purified graph} at the previous time step $t - 1$. We first add all incoming nodes and edges (see Sec.~\ref{sec:prelim}) to obtain the \textit{unpurified graph} $\hat{G}^{(t)} := \tilde{G}^{(t - 1)} \cup \Delta G^{(t)}$, and then we identify noise edges among the incoming edges $\Delta E^{(t)}$.\footnote{For $t=1$, we see $\tilde{G}^{(1)}$ as an initial graph $G^{(1)}$.}
Here, we only consider $\Delta E^{(t)}$ as candidate edges for efficiency since noise edges in $\tilde{G}^{(t - 1)}$ have already been purified in previous time steps.
Each candidate edge $(v_i, v_j) \in \Delta E^{(t)}$ is given a \textit{purification score} $S^{(t)}(v_i, v_j)$, and the bottom-$K$ edges w.r.t. the scores are removed.

\subsection{Long-term Module $M_{L}$}
\label{sec:method:long_module}

The long-term module $M_{L}$ captures temporal dependencies over extended periods (e.g., contextual patterns) and detects noise-like edges that deviate from these patterns, by using self-attention~\cite{vaswani2017attention}.

At each time step $t$, we first extract intrinsic information from the unpurified graph $\hat{G}^{(t)}$ and node attributes $X^{(t)}$ by computing the node embedding matrix $\mathbf{H}^{(t)}$ using a graph convolutional network (GCN) $f_L^{(t)}$
i.e., $\mathbf{H}^{(t)} = f_L^{(t)}(\hat{G}^{(t)}, \mathbf{X}^{(t)})$.\footnote{We use GCNs as a natural choice for graphs, while any suitable model can be used.}
To incorporate long-term temporal dependencies, for each node, we consider the embeddings from all the previous time steps.
For each node $v_i\in V^{(t)}$, let $\vech^{(t)}_i$ denote its embedding at time step $t$, and let $\mathcal{M}^{(t)}_{i}$ denote the memory set of its past embeddings, i.e.,
$\mathcal{M}^{(t)}_{i}=\{\vech^{(t)}_{i}, \vech^{(t-1)}_{i}, \cdots, \vech^{(k_{i})}_{i}\}$, where $k_{i}$ is the time step when $v_{i}$ first appears.
A self-attention mechanism, parameterized by $\mathbf{W}_{Q}$, $\mathbf{W}_{K}$, and $\mathbf{W}_{V}$, fuses these embeddings to produce a refined representation that captures long-term evolving structural patterns:
    $\vecz^{(t)}_{i} = \sum\nolimits_{\vech^{(\tau)}_{i} \in \mathcal{M}^{(t)}_{i}} \alpha^{(\tau)}_{i} \mathbf{W}_{V} \vech^{(\tau)}_{i}$
with
    $\alpha^{(\tau)}_{i} = \frac{\exp\left((\mathbf{W}_{Q} \vech^{(t)}_{i})^{T}(\mathbf{W}_{K} \vech^{(\tau)}_{i})\right)}{\sum\nolimits_{\vech^{(\tau')}_{i} \in \mathcal{M}^{(\tau')}_{i}} \exp\left((\mathbf{W}_{Q} \vech^{(t)}_{i})^{\top}(\mathbf{W}_{K} \vech^{(\tau')}_{i})\right)}$.

We then measure how much each newly added edge aligns with long-term contextual patterns.
For each edge $(v_i,v_j)$, we compute its sub-score $S^{(t)}_{L}(v_{i}, v_{j}) = \sigma(\vecz^{(t)}_{i}\mathbf{W}_{L}\vecz^{(t)}_{j} + \vecb_{L})$ by feeding the pair $(\vecz^{(t)}_{i}, \vecz^{(t)}_{j})$ into a bilinear layer parameterized by $\mathbf{W}_L$ and $\vecb_L$, followed by a sigmoid function $\sigma(\cdot)$.
Notably, $\mathbf{W}_L$ is constructed as a symmetric matrix to ensure order invariance.
Noise-like edges that significantly deviate from long-term contextual patterns are expected to have low sub-scores from $M_{L}$.

{
\begin{figure}[t]
\centering
\scalebox{1.00}{
    \begin{minipage}{\linewidth}
    \begin{algorithm}[H]
    \caption{Consistency score}\label{alg:ood}
    \KwIn{
    (1) an edge $(v_i,v_j)$, (2) a time step $t$, (3) a surrogate GCN model $f_{S}^{(t)}$, (4) an unpurified graph $\hat{G}^{(t)}$, (5) a node features $\mathbf{X}^{(t)}$.
    }
    \KwOut{
    consistency score of the given edge $(v_i,v_j)$.
    }
    $\mathbf{L}^{(t)} \gets f_{S}^{(t)}(\hat{G}^{(t)}, \mathbf{X}^{(t)})$\;
    $\textit{K}_{i} \gets \{\textit{KL-divergence}(\vecl^{(t)}_{i}, \vecl^{(t)}_{k})| v_k \in N^{(t)}_{i} \}$\;
    $\mu_{i}, \sigma_{i} \gets mean(\textit{K}_{i}), stdev(\textit{K}_{i})$\;
    $Z^{(t)}_{i} \gets \frac{|\textit{KL-divergence}(\vecl^{(t)}_{i}, \vecl^{(t)}_{j}) - \mu_{i}|}{\sigma_{i}} $\;
    Repeat Lines 2-4 for the node $j$ to obtain $Z^{(t)}_{j}$\;
    \Return $-\frac{Z^{(t)}_{i} + Z^{(t)}_{j}}{2}$
    \end{algorithm}
    \end{minipage}
}
\end{figure}
}

\subsection{Short-term Module $M_{S}$}
\label{sec:method:short_module}

The short-term module $M_{S}$ captures short-term temporal patterns (e.g., consistency across consecutive time steps) and detects noise-like edges that deviate from these patterns, by computing statistical distances. 

To achieve this, we define a consistency score (Algorithm~\ref{alg:ood}) that quantifies how well each incoming edge aligns with existing edges.
This consistency score is primarily computed based on latent vectors derived from a surrogate GCN model $f_{S}^{(t)}$.
Specifically, at each time step $t$, the surrogate GCN model $f_{S}^{(t)}$ from the previous time step is applied to the unpurified graph $\hat{G}^{(t)}$ and node attributes $X^{(t)}$ to generate a latent matrix $\mathbf{L}^{(t)}=f_{S}^{(t)}(\hat{G}^{(t)}, \mathbf{X}^{(t)})$ consisting of latent vectors $\vecl_i^{(t)}$ for all the nodes $v_i$ in $\hat{G}^{(t)}$.
\footnote{We assume the surrogate GCN model $f_{S}^{(t)}$ is trained at the previous time step for some downstream task, e.g., node classification.
The details of $f_{S}^{(t)}$'s appear in Sec.~\ref{sec:exp}.}
Using these latent vectors, given an edge $(v_i, v_j)$, we first evaluate the latent-vector-based statistics from $v_i$'s perspective by statistically comparing $v_j$'s latent vector with the latent vectors of $v_i$'s neighbors in $N^{(t)}_{i}$.
Specifically, we measure KL divergence (Line 2) and then perform a Z-score test by computing the mean $\mu^{(t)}_{i}$ and standard deviation $\sigma^{(t)}_{i}$ of the KL divergences (Line 3) to obtain the Z-score $Z^{(t)}_{i}$ (Line 4).
We repeat the same procedure for $v_j$ (Line 5).
Finally, we take the average of those two Z-scores and attach $(-)$ sign to define the \textit{consistency score} of the edge (Line 6).

The sub-score of $M_{S}$ for the edge $(v_{i}, v_{j})$, denoted as $S^{(t)}_{S}(v_{i}, v_{j})$, is simply this consistency score.
Noise-like edges, which significantly deviate from short-term consistency patterns, are expected to have low sub-scores from $M_{S}$.

\subsection{Ensemble Module}
\label{sec:method:ensemble}
We employ an ensemble approach to combine the sub-scores from the modules mentioned above.
To address the challenge of capturing temporal patterns during the early stages without enough training data accumulated, we incorporate an additional proximity-based sub-score to complement these modules.

The weights for the sub-scores from modules $M_{L}$ and $M_{S}$ are derived from the corresponding representations/latent vectors, ${\vecz}_i^{(t)}$'s and ${\vecl}_i^{(t)}$'s.
For each sub-score $S_L^{(t)}(v_i, v_j)$ from $M_{L}$, the element-wise mean and max of the representations are concatenated:
$\mathbf{x}^{(t)}_{L}(v_i,v_j) = [\text{elementMean}(\vecz^{(t)}_{i}, \vecz^{(t)}_{j}) || \text{elementMax}(\vecz^{(t)}_{i}, \vecz^{(t)}_{j})]$, 
where $||$ denotes vector concatenation.
The vector $\mathbf{x}^{(t)}_{L}(v_i,v_j)$ is then passed through a two-layer MLP to compute the weight:
$a^{(t)}_{L}(v_i,v_j) = \text{MLP}_{L}(\mathbf{x}^{(t)}_{L}(v_i,v_j))$. 
Similarly, the weight $a^{(t)}_{S}(v_i,v_j)$ for each sub-score $S_S^{(t)}(v_i, v_j)$ from $M_{S}$ is computed using the latent vectors ${\vecl}_i^{(t)}$'s.
For the proximity-based sub-scores $S^{(t)}_{P}(v_i, v_j)$'s, their weight is a hyperparameter $w_p$ shared for all edges.

We normalize the weights using softmax, i.e.,
$\hat\alpha^{(t)}_{L}(v_i,v_j), \hat\alpha^{(t)}_{S}(v_i,v_j), \hat\alpha^{(t)}_{P}(v_i,v_j) = \text{softmax}(a^{(t)}_{L}(v_i,v_j), a^{(t)}_{S}(v_i,v_j), w_p)$, and normalize the sub-scores $S^{(t)}_{S}$ and $S^{(t)}_{P}$ to $[0, 1]$ using min-max normalization.
For each edge $(v_i, v_j)$, the \textit{final purification score} $S^{(t)}(v_i,v_j)$ is the weighted sum of the normalized sub-scores:
$S^{(t)}(v_i,v_j) = \hat\alpha^{(t)}_{L}(v_i,v_j) S^{(t)}_{L}(v_i,v_j) + \hat\alpha^{(t)}_{S}(v_i,v_j) S^{(t)}_{S}(v_i,v_j) + \hat\alpha^{(t)}_{P}(v_i,v_j) S^{(t)}_{P}(v_i,v_j)$.

\subsection{Self-supervised Training Procedure}
\label{sec:method:training}

The proposed method, \algname,  employs self-supervised learning to identify noise edges without any ground-truth labels on noise edges, by assigning pseudo-labels to node pairs in the unpurified graph.

At each time step $t$, positive pseudo-labels are assigned to all the edges in the unpurified graph $\hat{G}^{(t)} = (\hat{V}^{(t)}, \hat{E}^{(t)})$, i.e., $T_p = \hat{E}^{(t)}$, and negative pseudo-labels to randomly selected non-edges, i.e., $T_n \subseteq \{(v_i,v_j) \mid (v_i,v_j) \notin \hat{E}^{(t)}\}$, with $|T_p| = |T_n|$ for balanced training.

Noise-like edges may also be assigned positive pseudo labels, potentially misleading the training process.
To mitigate this, we filter out low-scoring positive pseudo-labels during training, keeping only the top $\beta\%$ of edges based on the final purification scores $S^{(t)}(v_i, v_j)$'s (Sec.~\ref{sec:method:ensemble}), where $\beta$ is a hyperparameter.
We compute binary cross-entropy loss on the filtered positive sample set $T_p(\beta)$: 
$\mathcal{L} =  -\frac{1}{|T_p(\beta)|}\sum_{(v_i,v_j) \in T_p(\beta)} \log(S^{(t)}(v_i,v_j)) -\frac{1}{|T_n|} \sum_{(v_i,v_j) \in T_n} \log(1-S^{(t)}(v_i,v_j))$.

\subsection{Complexity Analysis}
\label{sec:method:complexity}
The time complexity of \algname involves inferring sub-scores from each module and combining them through the ensemble model.
At time step $t$, consider the unpurified graph $\hat{G}^{(t)}$ with $n$ nodes and $m$ edges.
Let $\bar{k}$ denote the average node degree, and $\Delta m$ be the number of incoming edges $\Delta E^{(t)}$. 
Let $d$ represent the dimension of the node features.
Assume that the GCNs have $l$ layers, each with hidden dimensions on the order of $d$.

The time complexity of \algname is dominated by three main operations: (1) GCN inference in both long-term and short-term modules, each with complexity $O(ld(nd + m))$; (2) the long-term self-attention mechanism, which processes one query against historical embeddings per node, with complexity $O(ntd^{2})$; and (3) computations for incoming edges, including statistical distances ($O(\Delta m \bar{k} d)$),  Adamic-Adar-based proximity sub-scores ($O(\Delta m \bar{k})$), and weights from a two-layer MLP ($O(\Delta m d^{2})$).
Combining these terms, the total complexity per time step is $O(ld(nd + m) + ntd^{2} + \Delta m (\bar{k}d + \bar{k} + d^{2}))$.

These results indicate that complexity of \algname per time step is linear with respect to the number of nodes $n$ and the number of edges $m$, and the empirical results in Online Appendix~\cite{code} further substantiate this analysis.

\section{Experiments}
\label{sec:exp}

In this section, we evaluate \algname to answer the Q1-Q3:
\begin{enumerate}
\item[Q1.] \textbf{Accuracy}: How accurately does \algname identify noise edges?
\item[Q2.] \textbf{Effectiveness}: How much does \algname enhance node classification?
\item[Q3.] \textbf{Ablation Study}: Does each component of \algname improve performance?
\end{enumerate}

\subsection{Experiment Settings}
\label{sec:exp:settings}
\smallsection{Datasets.} 
We use five real-world datasets, comprising  (1) 
three citation graphs: Aminer, Patent, and arXivAI, (2) a friendship graph: School,
and (3) a co-authorship graph: DBLP. 
Some basic statistics of the datasets and details about data pre-processing are provided in Online Appendix~\cite{code}.

\smallsection{Noisy time-evolving graph generation.} 
For numerical evaluation, we introduce noise edges that are not present in the original graph into each dataset. 
Specifically, at each time step $t$, we add noise edges corresponding to $30\%$ of the newly added edges $\Delta{E}^{(t)}$ to the graph. 
These noise edges are sampled uniformly at random from non-adjacent node pairs belonging to different classes.
For each dataset, we generate ten noisy graphs, and all results are averaged over them.

\smallsection{Competitors.} 
We evaluate \algname against six competitors, comprising 
(1) two local similarity-based methods: Adamic-Adar index (\adamic) and  Jaccard coefficient (\jaccard), 
(2) three global similarity-based methods: \svd, \gdc~\cite{klicpera2019diffusion}, and \tiara~\cite{lee2023time}\footnote{We adapt \gdc~\cite{klicpera2019diffusion} and \tiara~\cite{lee2023time}, originally designed for edge augmentation, for graph purification by removing edges with the least similarity or proximity.}, and
(3) a GNN-based approach: \leo~\cite{jo2025measuring}.
Details of the competitors can be found in Online Appendix~\cite{code}.

\smallsection{Node classification.}
In Sec.~\ref{sec:exp:effectiveness},
to evaluate effectiveness, we compare node classification performance on graphs purified by \algname and its competitors. 
Specifically, at each time step, a graph convolutional network (GCN) is trained on the purified graph.
It is also used as the surrogate GCN model $f_S^{(t)}$ in the short-term module $M_S$ (see Sec.~\ref{sec:method:short_module}).\footnote{Specifically, the surrogate GCN model $f_S^{(t)}$ in the short-term module is the GCN that is trained at time step $t-1$ for node classification.}
The nodes are randomly divided into training, validation, and test sets in a 1:1:8 ratio, with the same split maintained across all time steps.

\smallsection{Hyperparameters.}  
For \algname, we select the hyperparameters $\beta$ and 
$w_p$ via grid search over $\{0.1, 0.2, 0.3\}$, and $\{1, 5, 10, 20\}$, respectively. 
The hidden dimension of each trainable parameter was set to 64. 
For all methods, we use the ground-truth purification budget $K$ for each graph at time step $t$ (i.e., $K = 0.3\times|\Delta E^{t}|$) to ensure a fair comparison.
Detailed search spaces and the selected hyperparameters for all methods are provided in Online Appendix~\cite{code}.

\begin{table}[p]
    \centering
    \caption[Accuracy.]{\label{tab:q1} 
    (Q1) \underline{\smash{Accuracy.}}
    \algname accurately identifies noise edges. Each entry in a table represents the proportion of noise removed by each method at each time step. Note that all methods are tested under the same purification budget.
    }
    \begin{subtable}{\textwidth}
        \centering
        \begin{tabular}{ c||c|c|c|c|c|c } 
            \hline
             & \textbf{Method} & $t=2$ & $t=4$ & $t=6$ & $t=8$ & $t=10$ \\
            \hline
            \hline
\multirow{8}{*}{\rotatebox[origin=c]{90}{School}} & \svd & 58.41$\pm$1.49 & 43.01$\pm$1.58 & 39.47$\pm$1.17 & 37.39$\pm$0.65 & 36.16$\pm$0.77\\
& \adamic & 73.19$\pm$1.83 & \underline{63.33$\pm$1.02} & \underline{60.41$\pm$0.97} & \underline{62.81$\pm$1.26} & 64.28$\pm$1.17\\
& \jaccard & \underline{73.26$\pm$1.85} & 63.12$\pm$1.08 & 60.29$\pm$1.24 & 62.68$\pm$1.18 & \underline{64.33$\pm$1.20}\\
& \gdc & 66.09$\pm$1.58 & 53.88$\pm$1.59 & 50.99$\pm$1.86 & 49.33$\pm$1.37 & 47.23$\pm$1.24\\
& \tiara & 62.10$\pm$2.43 & 45.36$\pm$1.86 & 38.99$\pm$1.37 & 38.73$\pm$1.34 & 38.00$\pm$1.24\\
& \leo & 72.32$\pm$2.40 & 60.69$\pm$1.30 & 57.22$\pm$1.34 & 60.29$\pm$1.25 & 61.49$\pm$0.84\\
\cline{2-7}
& \multirow{2}{*}{\algname} & \textbf{76.38$\pm$1.95} & \textbf{69.46$\pm$0.89} & \textbf{66.57$\pm$0.79} & \textbf{69.17$\pm$1.05} & \textbf{69.84$\pm$0.91}\\
& & \textbf{(+4.3\%)} & \textbf{(+9.7\%)} & \textbf{(+10.2\%)} & \textbf{(+10.1\%)} & \textbf{(+8.6\%)}\\
            \hline
\multirow{8}{*}{\rotatebox[origin=c]{90}{Patent}} & \svd & 90.67$\pm$0.42 & 86.86$\pm$0.42 & 84.47$\pm$0.28 & 81.52$\pm$0.46 & 80.16$\pm$0.52\\
& \adamic & 61.25$\pm$0.69 & 43.02$\pm$0.52 & 36.10$\pm$0.53 & 33.08$\pm$0.38 & 31.12$\pm$0.37\\
& \jaccard & 61.25$\pm$0.69 & 43.02$\pm$0.52 & 36.10$\pm$0.53 & 33.08$\pm$0.38 & 31.12$\pm$0.37\\
& \gdc & 88.91$\pm$1.60 & 84.41$\pm$1.06 & 82.87$\pm$0.84 & 82.42$\pm$0.75 & 80.44$\pm$0.80\\
& \tiara & 54.31$\pm$0.38 & 31.23$\pm$0.33 & 23.61$\pm$0.36 & 20.20$\pm$0.33 & 18.74$\pm$0.29\\
& \leo & \underline{91.87$\pm$0.87} & \underline{89.67$\pm$0.46} & \underline{88.74$\pm$0.80} & \underline{88.30$\pm$0.64} & \underline{87.48$\pm$0.64}\\
\cline{2-7}
& \multirow{2}{*}{\algname} & \textbf{92.25$\pm$0.72} & \textbf{89.88$\pm$0.47} & \textbf{89.09$\pm$0.33} & \textbf{88.61$\pm$0.28} & \textbf{87.79$\pm$0.41}\\
& & \textbf{(+0.4\%)} & \textbf{(+0.2\%)} & \textbf{(+0.4\%)} & \textbf{(+0.4\%)} & \textbf{(+0.4\%)}\\
            \hline
\multirow{8}{*}{\rotatebox[origin=c]{90}{DBLP}} & \svd & 62.56$\pm$0.12 & 44.50$\pm$0.22 & 39.19$\pm$0.13 & 37.15$\pm$0.18 & 36.35$\pm$0.15\\
& \adamic & \underline{89.87$\pm$0.10} & 84.61$\pm$0.13 & 84.08$\pm$0.11 & 84.86$\pm$0.09 & 85.70$\pm$0.08\\
& \jaccard & 89.87$\pm$0.11 & \underline{84.68$\pm$0.15} & \underline{84.15$\pm$0.14} & \underline{84.90$\pm$0.12} & \underline{85.73$\pm$0.10}\\
& \gdc & O.O.T. & O.O.T. & O.O.T. & O.O.T. & O.O.T.\\
& \tiara & O.O.T. & O.O.T. & O.O.T. & O.O.T. & O.O.T.\\
& \leo & 89.29$\pm$0.32 & 83.12$\pm$0.47 & 81.56$\pm$0.41 & 81.28$\pm$0.41 & 81.28$\pm$0.33\\
\cline{2-7}
& \multirow{2}{*}{\algname} & \textbf{92.34$\pm$0.12} & \textbf{88.40$\pm$0.26} & \textbf{87.85$\pm$0.25} & \textbf{88.21$\pm$0.25} & \textbf{88.68$\pm$0.23}\\
& & \textbf{(+2.7\%)} & \textbf{(+4.4\%)} & \textbf{(+4.4\%)} & \textbf{(+3.9\%)} & \textbf{(+3.4\%)}\\
            \hline
\multirow{8}{*}{\rotatebox[origin=c]{90}{ArXivAI}} & \svd & 73.28$\pm$0.09 & 60.54$\pm$0.07 & 56.44$\pm$0.07 & 54.98$\pm$0.06 & 53.85$\pm$0.05\\
& \adamic & 81.07$\pm$0.07 & 71.66$\pm$0.08 & 69.39$\pm$0.06 & 68.72$\pm$0.06 & 68.50$\pm$0.06\\
& \jaccard & 81.11$\pm$0.06 & 71.71$\pm$0.08 & 69.45$\pm$0.07 & 68.76$\pm$0.07 & \underline{68.53$\pm$0.07}\\
& \gdc & O.O.T. & O.O.T. & O.O.T. & O.O.T. & O.O.T.\\
& \tiara & O.O.T. & O.O.T. & O.O.T. & O.O.T. & O.O.T.\\
& \leo & \underline{83.81$\pm$0.35} & \underline{74.39$\pm$0.22} & \underline{70.87$\pm$0.23} & \underline{68.99$\pm$0.28} & 67.86$\pm$0.24\\
\cline{2-7}
& \multirow{2}{*}{\algname} & \textbf{84.40$\pm$0.32} & \textbf{76.24$\pm$0.24} & \textbf{73.88$\pm$0.24} & \textbf{72.99$\pm$0.26} & \textbf{72.61$\pm$0.25}\\
& & \textbf{(+0.7\%)} & \textbf{(+2.5\%)} & \textbf{(+4.2\%)} & \textbf{(+5.8\%)} & \textbf{(+6.0\%)}\\
            \hline
        \end{tabular}
    \end{subtable}

    \begin{subtable}{\textwidth}
        \centering
        \begin{tabular}{ c||c|c|c|c|c|c } 
            \hline
             & \textbf{Method} & $t=3$ & $t=4$ & $t=5$ & $t=6$ & $t=7$ \\
            \hline
            \hline
\multirow{8}{*}{\rotatebox[origin=c]{90}{Aminer}} & \svd & 60.87$\pm$0.55 & 49.75$\pm$0.54 & 41.99$\pm$0.61 & 37.54$\pm$0.69 & 34.67$\pm$0.51\\
& \adamic & 67.29$\pm$0.94 & 59.46$\pm$0.88 & 54.43$\pm$0.81 & 52.37$\pm$0.79 & 50.44$\pm$0.62\\
& \jaccard & 67.29$\pm$0.94 & 59.43$\pm$0.84 & 54.43$\pm$0.81 & 52.32$\pm$0.79 & 50.43$\pm$0.64\\
& \gdc & 64.33$\pm$1.02 & 55.83$\pm$1.10 & 49.80$\pm$0.97 & 45.87$\pm$0.91 & 43.89$\pm$0.84\\
& \tiara & 63.00$\pm$0.85 & 51.66$\pm$0.71 & 44.73$\pm$0.66 & 41.00$\pm$0.57 & 39.13$\pm$0.57\\
& \leo & \underline{82.70$\pm$0.88} & \underline{78.50$\pm$0.90} & \underline{75.55$\pm$0.84} & \underline{74.07$\pm$0.88} & \underline{73.40$\pm$0.89}\\
\cline{2-7}
& \multirow{2}{*}{\algname} & \textbf{84.21$\pm$1.27} & \textbf{80.86$\pm$1.53} & \textbf{79.17$\pm$1.30} & \textbf{78.74$\pm$1.01} & \textbf{78.28$\pm$0.98}\\
& & \textbf{(+1.8\%)} & \textbf{(+3.0\%)} & \textbf{(+4.8\%)} & \textbf{(+6.3\%)} & \textbf{(+6.6\%)}\\
            \hline
        \end{tabular}
    \end{subtable}

\end{table}
\begin{figure}[ht!]
    \centering \caption[Effectiveness.]{\label{tab:q2} (Q2) \underline{\smash{Effectiveness.}}
        \algname consistently enhances node classification performance, outperforming its competitors in most cases.
        On the DBLP and ArXivAI datasets, \gdc and \tiara exceed the six-hour time limit.
    }
    \includegraphics[width=\linewidth]{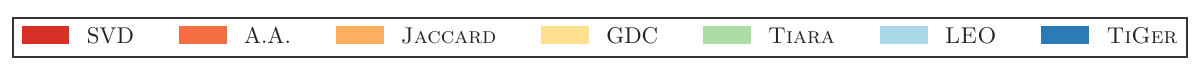} 
    \begin{subtable}{\linewidth}
        \centering 
        \begin{tabular}{c|c|c|c|c}
            \hline
            $t=2$ & $t=4$ & $t=6$ & $t=8$ & $t=10$ \\
            \hline\hline
            \includegraphics[width=0.186\linewidth]{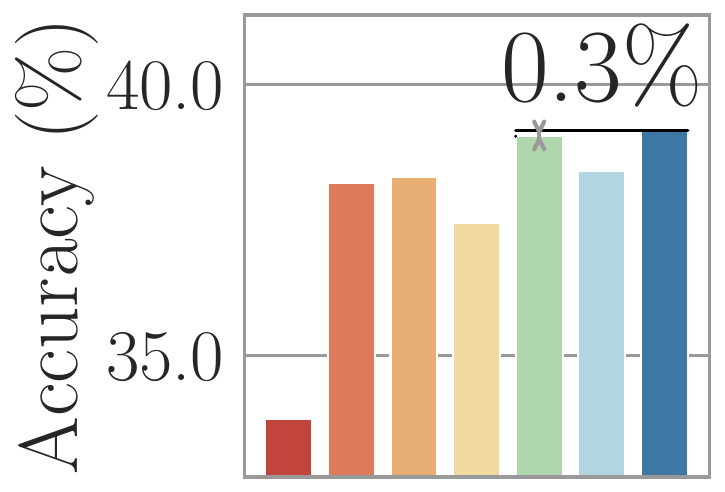} & 
            \includegraphics[width=0.186\linewidth]{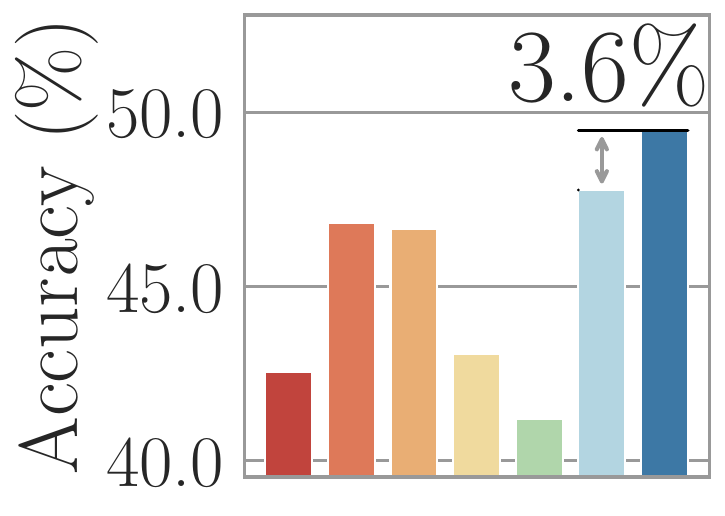} & 
            \includegraphics[width=0.186\linewidth]{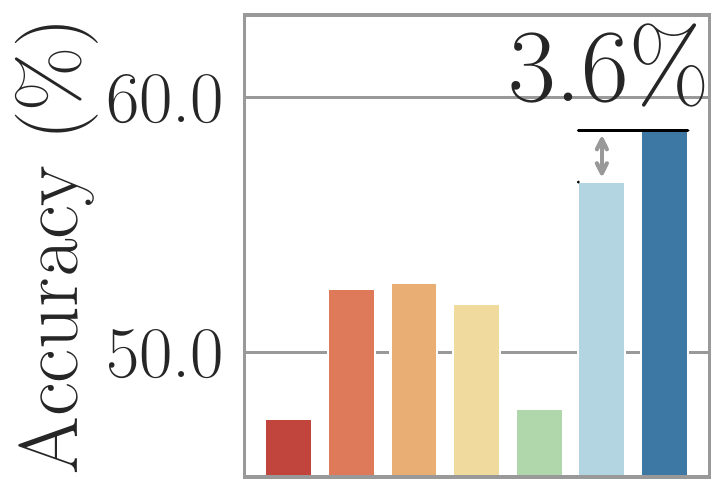} & 
            \includegraphics[width=0.186\linewidth]{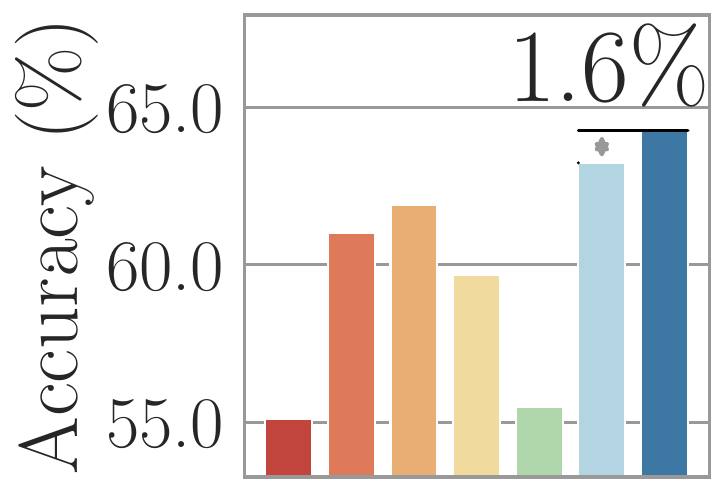} &
            \includegraphics[width=0.186\linewidth]{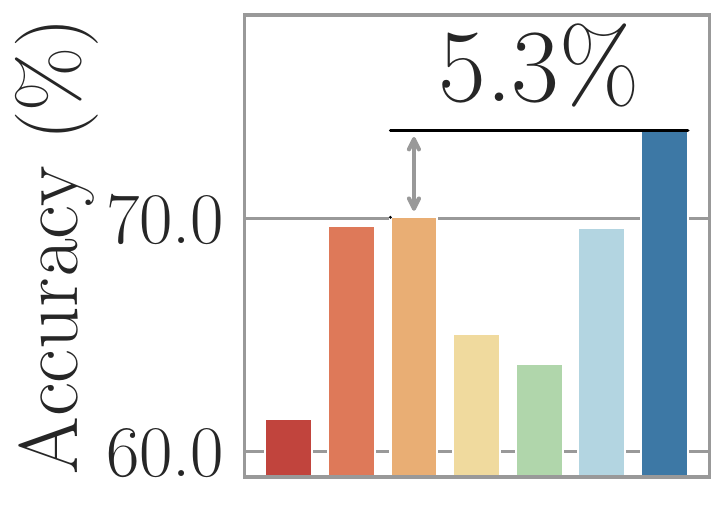} \\
            \hline
        \end{tabular}
        \caption{School}\label{tab:q2_school}
    \end{subtable}

    \begin{subtable}{\linewidth}
        \centering 
        \begin{tabular}{c|c|c|c|c}
            \hline
            $t=3$ & $t=4$ & $t=5$ & $t=6$ & $t=7$ \\
            \hline
            \includegraphics[width=0.186\linewidth]{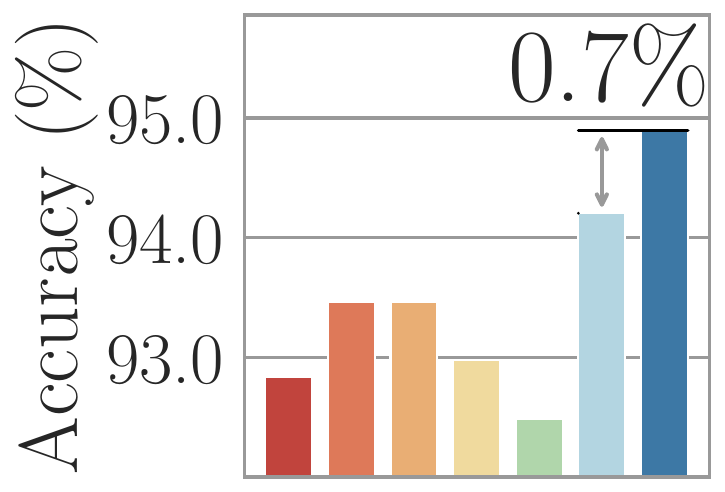} & 
            \includegraphics[width=0.186\linewidth]{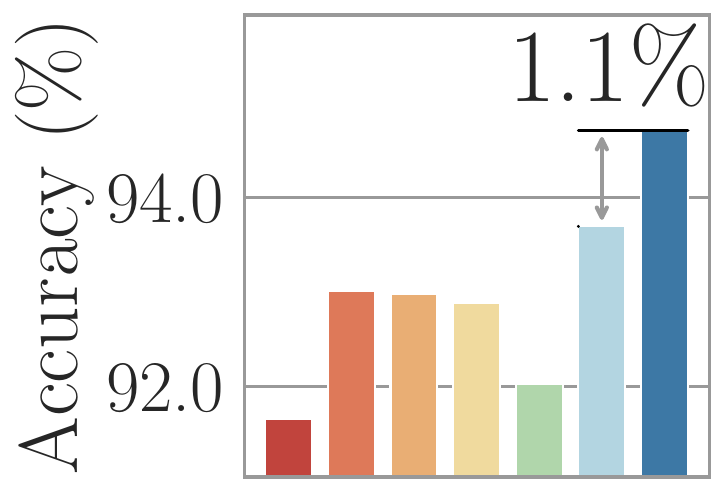} & 
            \includegraphics[width=0.186\linewidth]{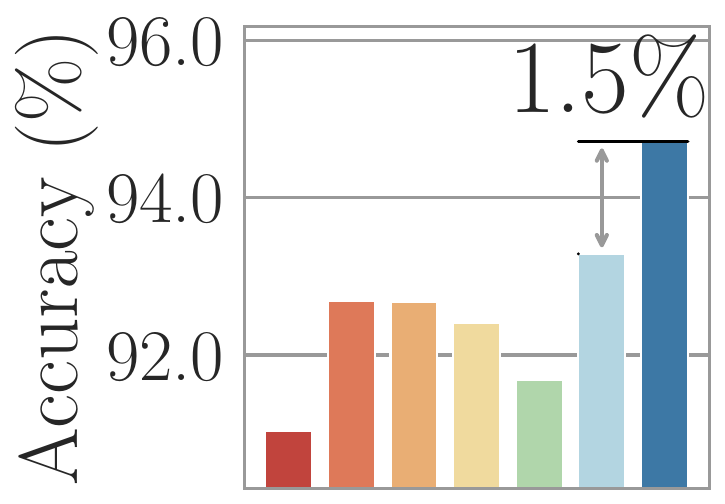} & 
            \includegraphics[width=0.186\linewidth]{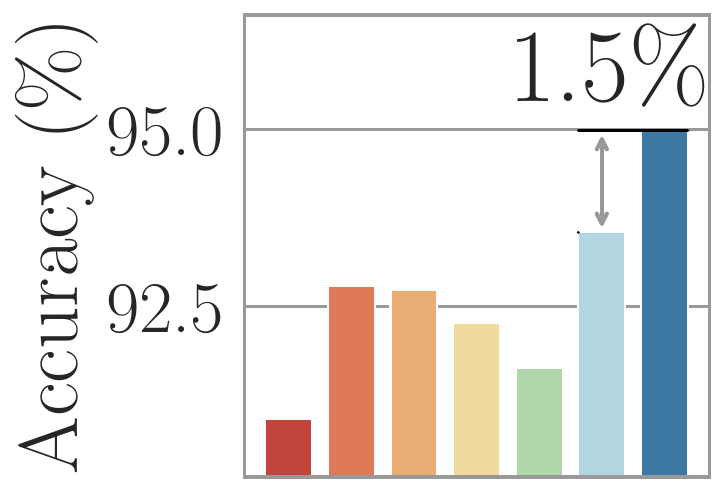} &
            \includegraphics[width=0.186\linewidth]{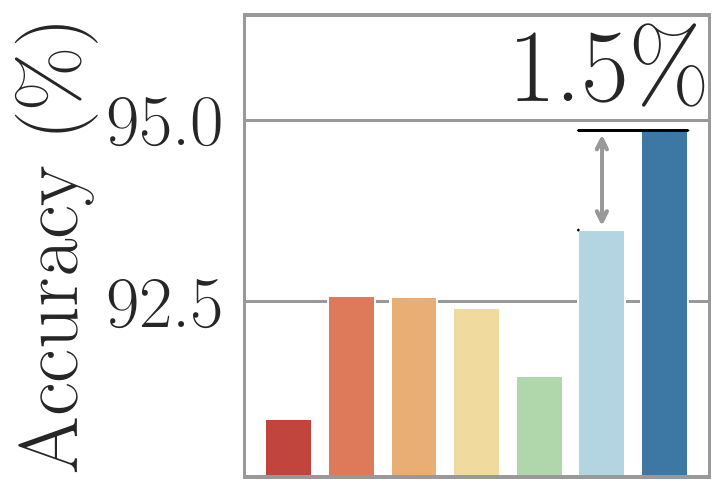} \\
            \hline
        \end{tabular}
        \caption{Aminer}\label{tab:q2_aminer}
    \end{subtable}

    \begin{subtable}{\linewidth}
        \centering 
        \begin{tabular}{c|c|c|c|c}
            \hline
            $t=2$ & $t=4$ & $t=6$ & $t=8$ & $t=10$ \\
            \hline\hline
            \includegraphics[width=0.186\linewidth]{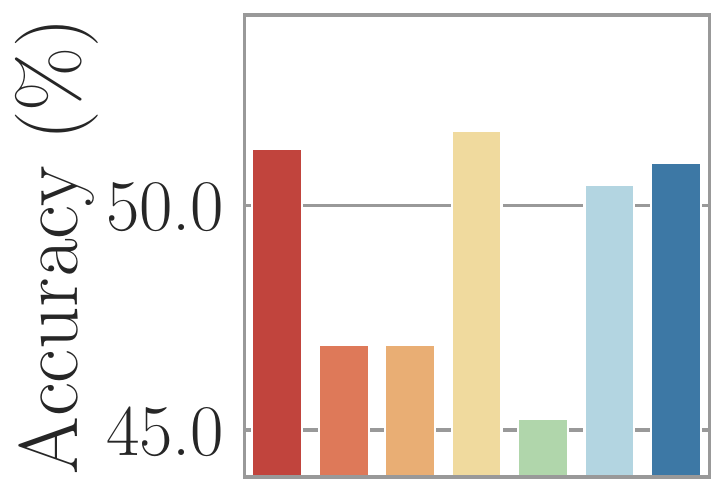} & 
            \includegraphics[width=0.186\linewidth]{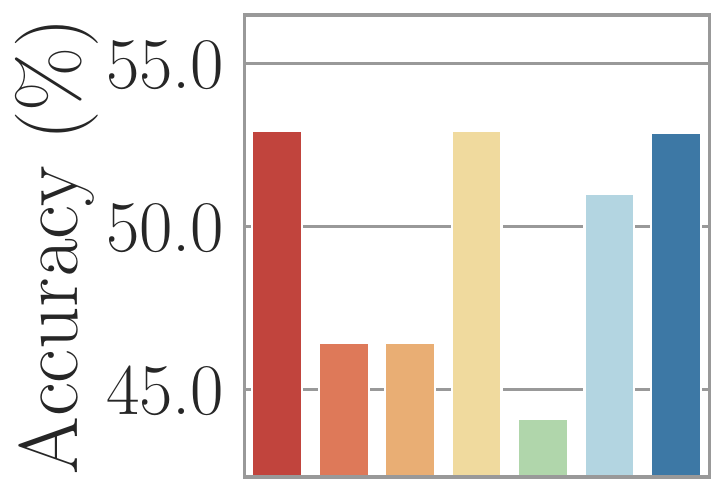} & 
            \includegraphics[width=0.186\linewidth]{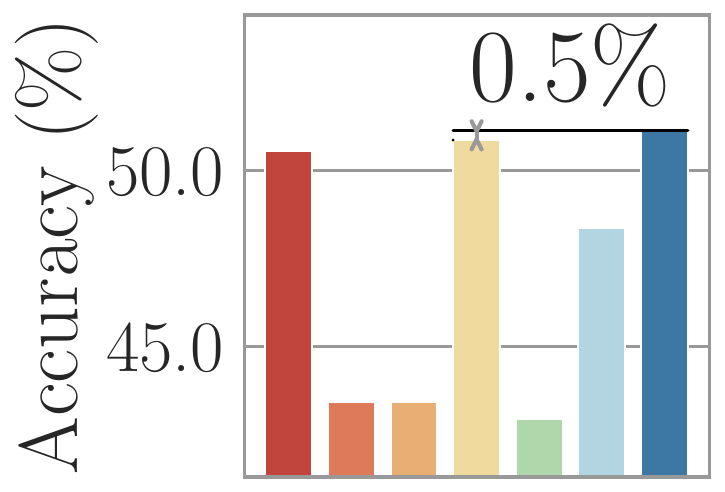} & 
            \includegraphics[width=0.186\linewidth]{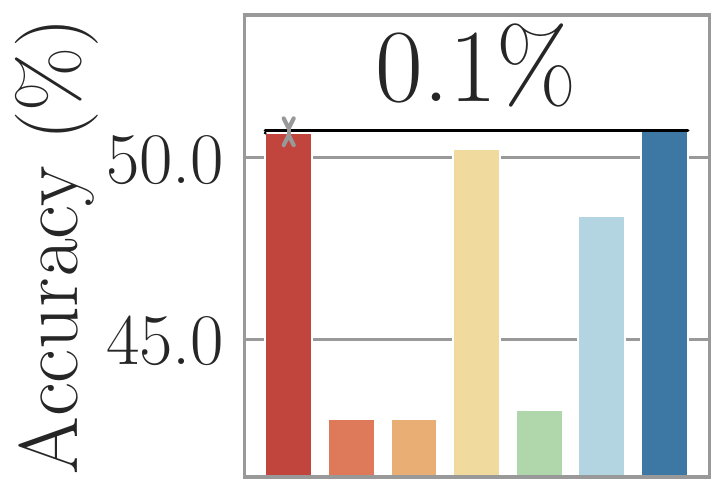} &
            \includegraphics[width=0.186\linewidth]{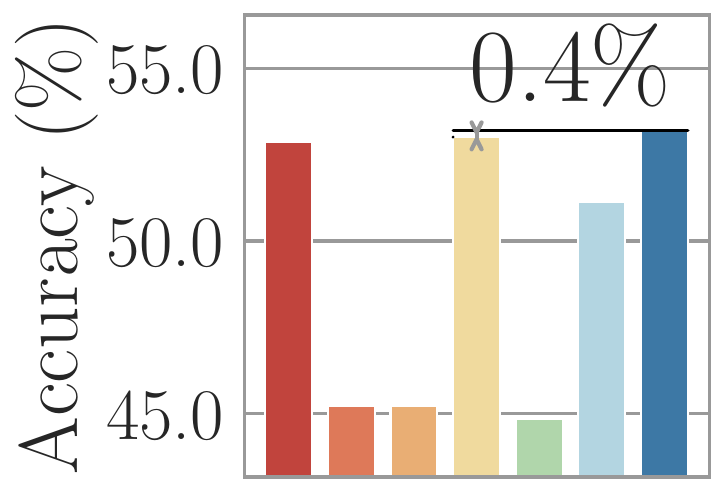} \\
            \hline
        \end{tabular}
        \caption{Patent}\label{tab:q2_patent}
    \end{subtable}
    
    \begin{subtable}{\linewidth}
        \centering 
        \begin{tabular}{c|c|c|c|c}
            \hline
            $t=2$ & $t=4$ & $t=6$ & $t=8$ & $t=10$ \\
            \hline\hline
            \includegraphics[width=0.186\linewidth]{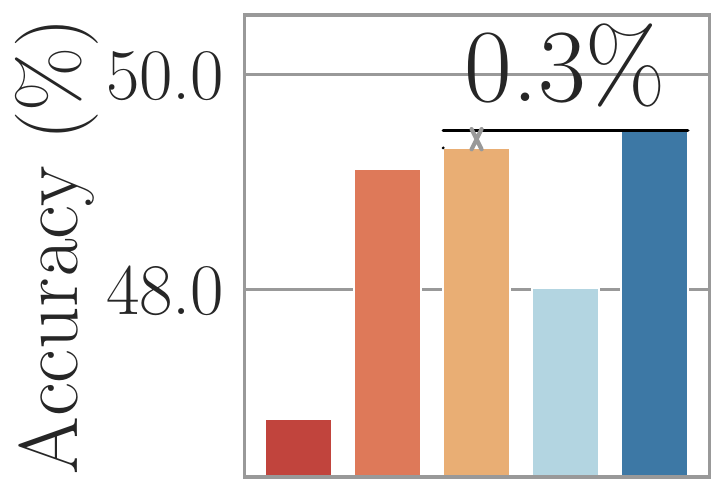} & 
            \includegraphics[width=0.186\linewidth]{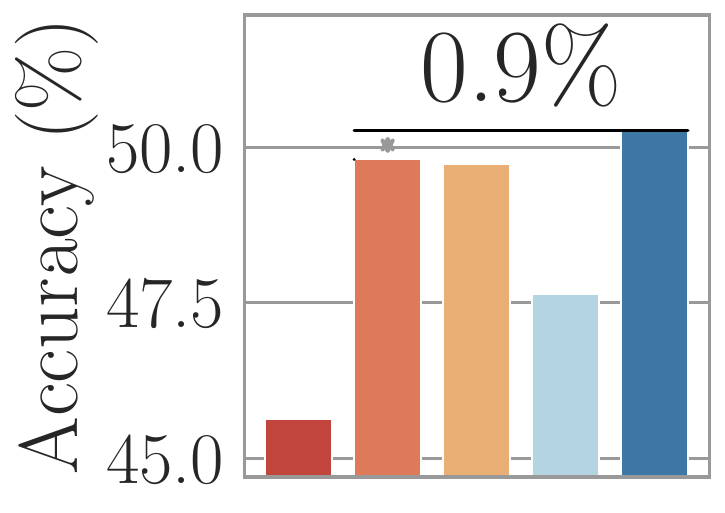} & 
            \includegraphics[width=0.186\linewidth]{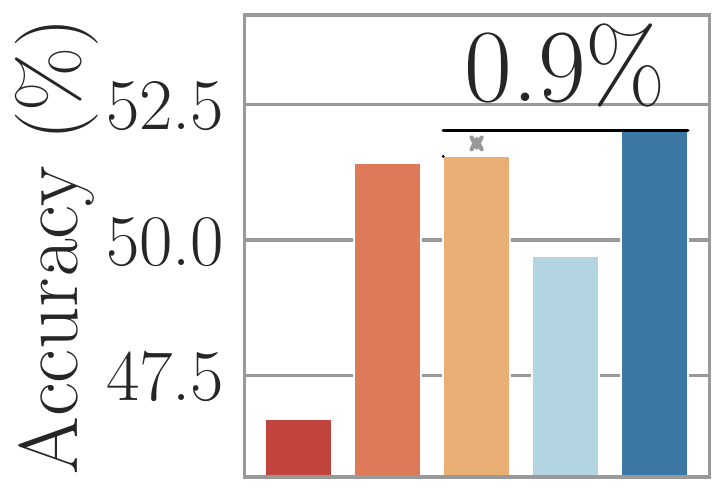} & 
            \includegraphics[width=0.186\linewidth]{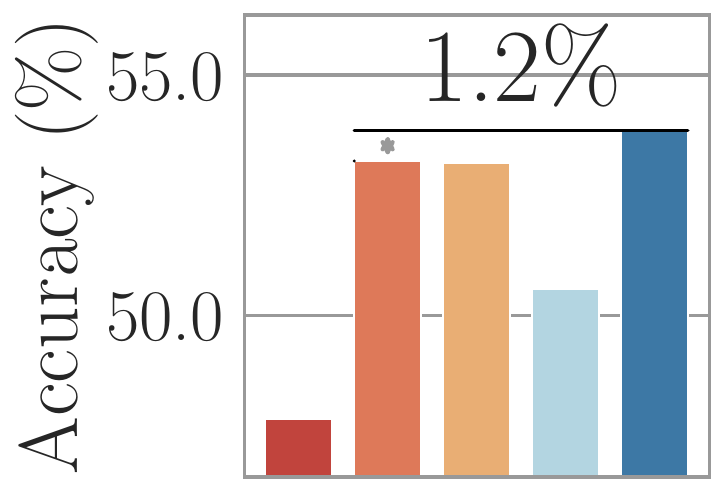} &
            \includegraphics[width=0.186\linewidth]{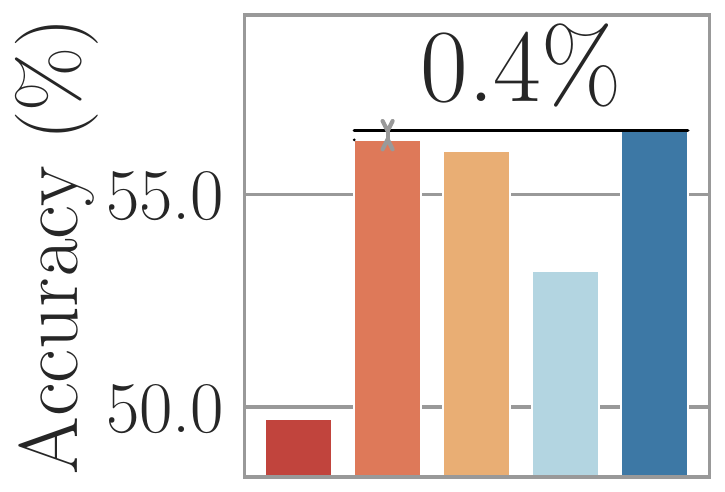} \\
            \hline
        \end{tabular}
        \caption{DBLP}\label{tab:q2_dblp}
    \end{subtable}
    
    \begin{subtable}{\linewidth}
        \centering 
        \begin{tabular}{c|c|c|c|c}
            \hline
            $t=2$ & $t=4$ & $t=6$ & $t=8$ & $t=10$ \\
            \hline\hline
            \includegraphics[width=0.186\linewidth]{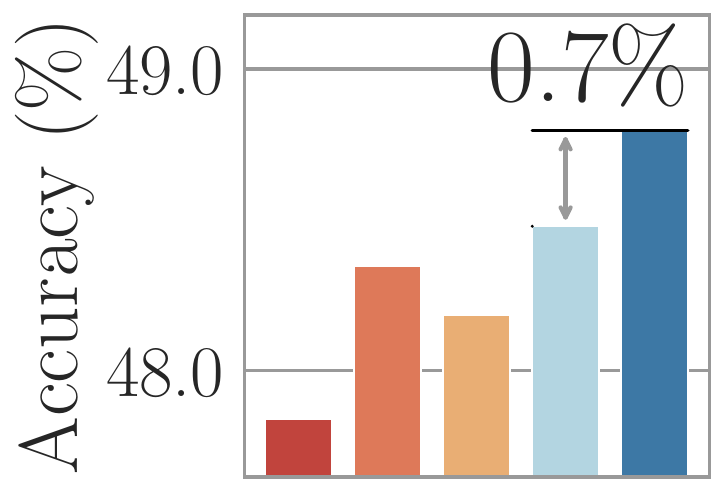} & 
            \includegraphics[width=0.186\linewidth]{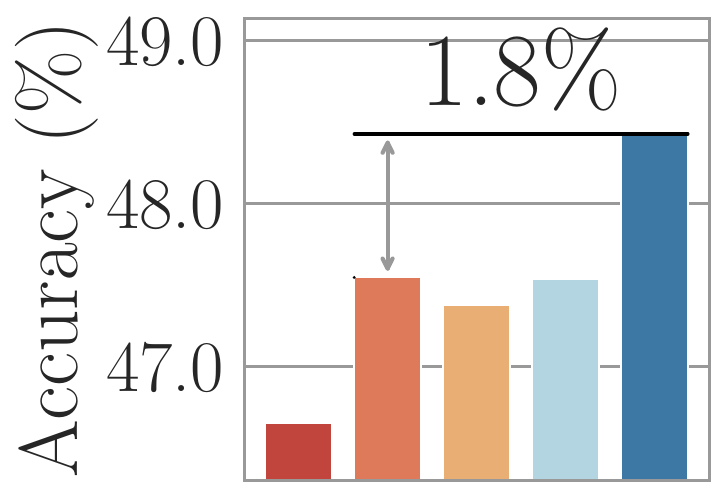} & 
            \includegraphics[width=0.186\linewidth]{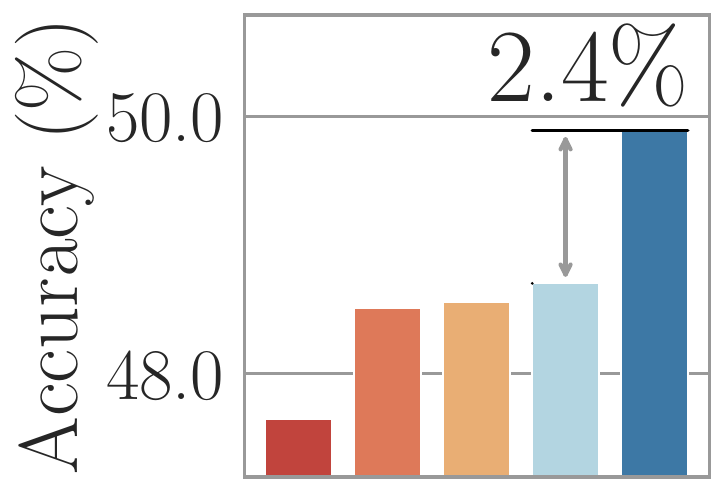} & 
            \includegraphics[width=0.186\linewidth]{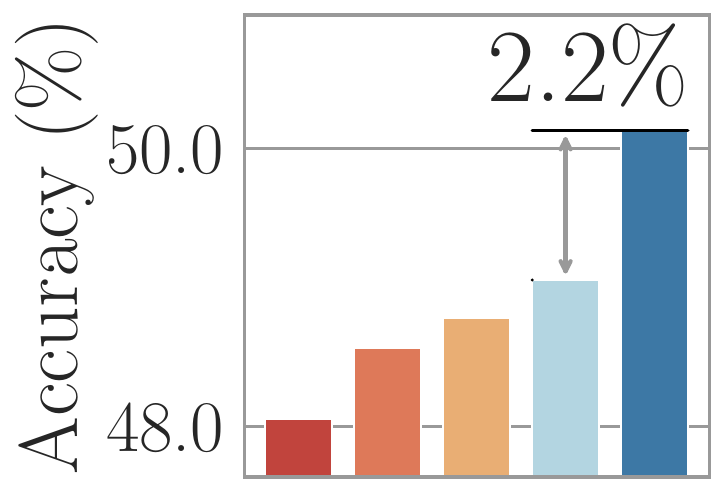} &
            \includegraphics[width=0.186\linewidth]{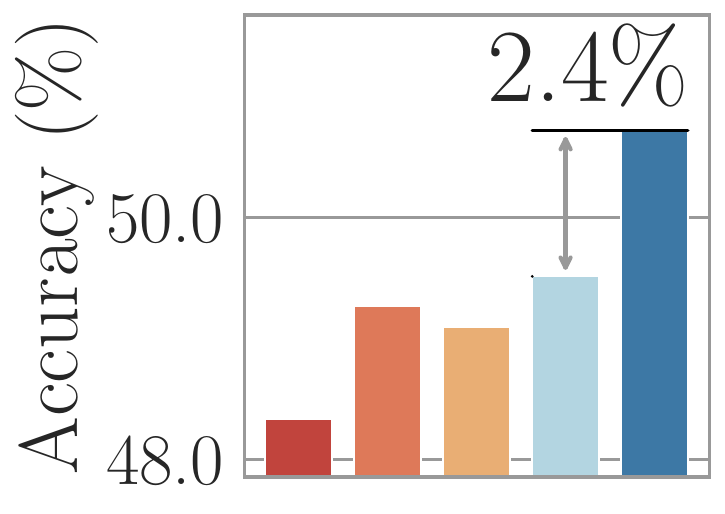} \\
            \hline
        \end{tabular}
        \caption{arXivAI\label{fig:q2_arxivai}}
    \end{subtable}
\end{figure}

\subsection{Q1. Accuracy} 
\label{sec:exp:accuracy}

In Table~\ref{tab:q1}, we report the proportion of noise edges removed by each method in each dataset and at each time step, under the same purification budget (i.e., the number of removed edges) for all methods.
Methods exceeding the six-hour time limit (e.g., TIARA and GDC on the DBLP and arXivAI datasets) are marked as \textit{out of time} (O.O.T.)
\algname consistently outperforms all its competitors in nearly all cases, with purification rates up to 10.2\% higher than the second-best method (at $t=6$ on School dataset).
These results show the effectiveness of \algname in removing noise edges, preserving the integrity of time-evolving graphs.

\subsection{Q2. Effectiveness} 
\label{sec:exp:effectiveness}

In Fig.~\ref{tab:q2}, we report the node classification accuracy at each time step on each time-evolving graph purified by each method (see Sec.~\ref{sec:exp:settings} for details on node classification). 
\algname consistently leads to performance improvements, ranking first in 23 out of 25 cases and second in the remaining two cases (at $t=2$ and $t=4$ on the Patent dataset).
Notably, using \algname for purification yields up to \textbf{$5.3\%$} higher node classification accuracy, compared to using the second-best purification method instead, as shown in Fig.~\ref{tab:q2}(a).

\begin{table}[t!]
    \centering
    \caption[Ablation Study.]{\label{tab:q3} 
    (Q3) \underline{\smash{Ablation Study.}}
     \algname achieves better purification accuracy than its variants, highlighting the complementary contributions of its long-term module ($M_L$) and its short-term module ($M_S$).
     Each entry represents the proportion of noise purified by each method at each time step in the ArXivAI dataset. 
    }
    \begin{subtable}{\textwidth}
        \centering
        \begin{tabular}{c|c||c|c|c|c|c } 
                \hline
                 \textbf{$M_{L}$} & \textbf{$M_{S}$} & $t=2$ & $t=4$ & $t=6$ & $t=8$ & $t=10$ \\
                \hline
                \hline
\xmark & \xmark & 83.73$\pm$0.27 & 74.51$\pm$0.36 & 71.07$\pm$0.26 & 69.35$\pm$0.34 & 68.29$\pm$0.33\\
\xmark & \cmark & 83.95$\pm$0.20 & 74.79$\pm$0.30 & 71.57$\pm$0.33 & 69.73$\pm$0.35 & 68.66$\pm$0.35\\
\cmark & \xmark & \underline{84.33$\pm$0.24} & \underline{76.16$\pm$0.20} & \underline{73.77$\pm$0.21} & \underline{72.64$\pm$0.22} & \underline{71.95$\pm$0.25}\\
\hline
\cmark & \cmark & \textbf{84.40$\pm$0.32} & \textbf{76.24$\pm$0.24} & \textbf{73.88$\pm$0.24} & \textbf{72.99$\pm$0.26} & \textbf{72.61$\pm$0.25}\\
            \hline
        \end{tabular}
    \end{subtable}
\end{table}

\subsection{Q3. Ablation Study} 
\label{sec:exp:ablation}

We evaluate the contributions of two key components of \algname, i.e., the long-term module $M_{L}$ and the short-term module $M_{S}$, through an ablation study. 
To isolate the contribution of $M_{L}$, we removed the self-attention mechanism, while the contribution of $M_{S}$ was evaluated by excluding the module entirely from the ensemble model.
We compare original \algname (utilizing both modules) with its three simplified variants (1) without both modules, (2) without $M_{L}$, and (3) without $M_{S}$.
As shown in Table~\ref{tab:q3}, where the ArXivAI dataset is used,  integrating both modules significantly improves noise identification accuracy.

\section{Conclusions}
\label{sec:conclusion}
In this paper, we proposed \algname, a self-supervised method for purifying time-evolving graphs. By leveraging dedicated modules to capture long-term patterns and short-term patterns, \algname identifies and filters out noise edges that deviate from these temporal patterns. Its proximity-based ensemble strategy further enhances robustness.
Our extensive experiments showed that \algname consistently outperforms its competitors in noise purification and downstream node classification. For reproducibility, we provide our code and datasets at~\cite{code}.

\subsubsection*{Acknowledgements.}
This work was supported by Institute of Information \& communications Technology Planning \& Evaluation (IITP) grant funded by the Korea government (MSIT) (No. RS-2024-00457882, AI Research Hub Project, 50\%)
(No. No. 2022-0-00157/RS-2022-II220157, Robust, Fair, Extensible Data-Centric Continual Learning, 40\%)
(RS-2019-II190075, Artificial Intelligence Graduate School Program (KAIST), 10\%).

%
%
%
%

\bibliographystyle{splncs04}
\bibliography{ref}

\end{document}